\pdfoutput=1

\documentclass[11pt]{article}

\usepackage[]{ACL2023}

\usepackage{times}
\usepackage{latexsym}

\usepackage[T1]{fontenc}

\usepackage[utf8]{inputenc}
\usepackage{graphicx}  
\usepackage{tabularx}
\usepackage{color} 
\usepackage{amsfonts}
\usepackage{amsmath}
\usepackage{multirow}
\usepackage{booktabs}
\usepackage{supertabular}

\usepackage{tabu}                     
\usepackage{multirow}                 
\usepackage{multicol}                 
\usepackage{multirow}                
\usepackage{float}                    
\usepackage{makecell}                 
\usepackage{booktabs}

\usepackage[ruled,linesnumbered]{algorithm2e}
\usepackage{microtype}

\usepackage{inconsolata}

\title{Prefix-diffusion: A Lightweight Diffusion Model for \protect\\ Diverse Image Captioning}

\author{Guisheng~Liu\textsuperscript{1}, Yi~Li\textsuperscript{1} , Zhengcong~Fei\textsuperscript{3}, Haiyan~Fu\textsuperscript{1},  Xiangyang~Luo\textsuperscript{2}, Yanqing~Guo\textsuperscript{1} \\
	\textsuperscript{1}Dalian University of Technology \\ 
	\textsuperscript{2}Information Engineering University, \textsuperscript{3}Meituan\\ 	 
	lgs0000@mail.dlut.edu.cn,~\{liyi, fuhy, guoyq\}@dlut.edu.cn \\
	xiangyangluo@126.com, feizhengcong@meituan.com}

\begin{document}
	\maketitle
	\begin{abstract}
		While impressive performance has been achieved in image captioning, the limited diversity of the generated captions and the large parameter scale remain major barriers to the real-word application of these systems. In this work, we propose a lightweight image captioning network in combination with continuous diffusion, called Prefix-diffusion. To achieve diversity, we design an efficient method that injects prefix image embeddings into the denoising process of the diffusion model. In order to reduce trainable parameters, we employ a pre-trained model to extract image features and further design an extra mapping network. Prefix-diffusion is able to generate diverse captions with relatively less parameters, while maintaining the fluency and relevance of the captions benefiting from the generative capabilities of the diffusion model. Our work paves the way for scaling up diffusion models for image captioning, and achieves promising performance compared with recent approaches.\footnote{Code will be released upon publication.}
	\end{abstract}

	\section{Introduction}
	
	Image captioning, which combines computer vision (CV) and natural language processing (NLP), focuses mainly on producing a description of an image. Existing works on image captioning typically employ an encoder-decoder architecture \citep{vinyals2015show, anderson2018bottom, zhou2020unified} to generate captions word-by-word. However, such models require large trainable parameters to bridge the visual and textual representations. By utilizing the powerful representation capability of pre-trained models like CLIP\citep{radford2021learning}, recent methods \citep{lovenia2022every, zhu2022visualize, mokady2021clipcap} map visual semantic information to language space for image captioning. Although autoregressive models have become the typical approach for image captioning, their left-to-right generative manner leads to cumulative errors. Moreover, human-like captions not only maintain fluency and relevance properties, but also contain diverse wordings and rich expressions.

	\begin{figure}[t]
		\begin{center}
			\includegraphics[width=1.0\linewidth]{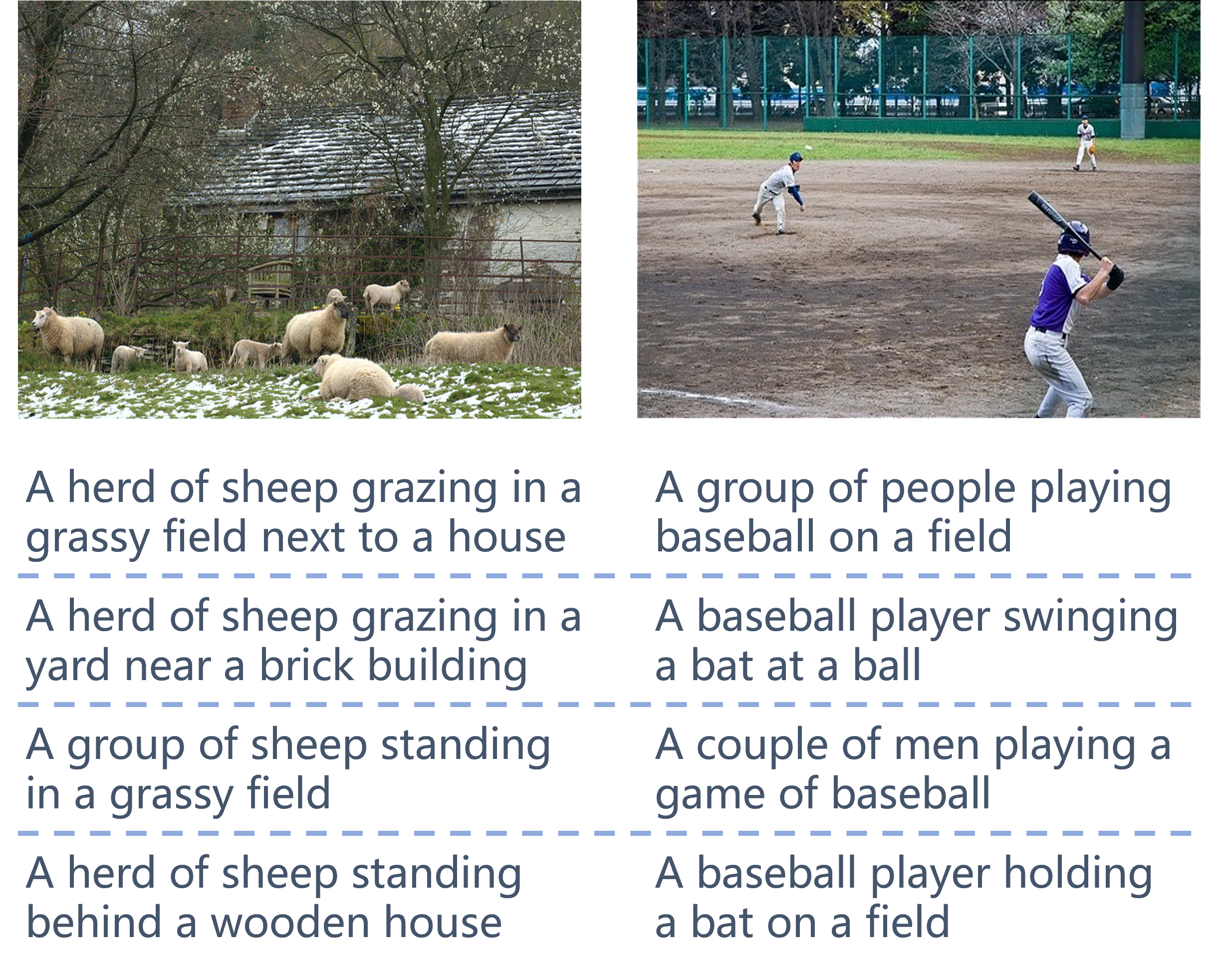}
		\end{center}
		\caption{The diverse captions generated by Prefix-diffusion. The model is trained on the COCO dataset. More examples will be given in the supplementary material.}
		\label{pic1}
	\end{figure}
	
	\begin{figure*}[t]
		\begin{center}
			\includegraphics[width=0.85\linewidth]{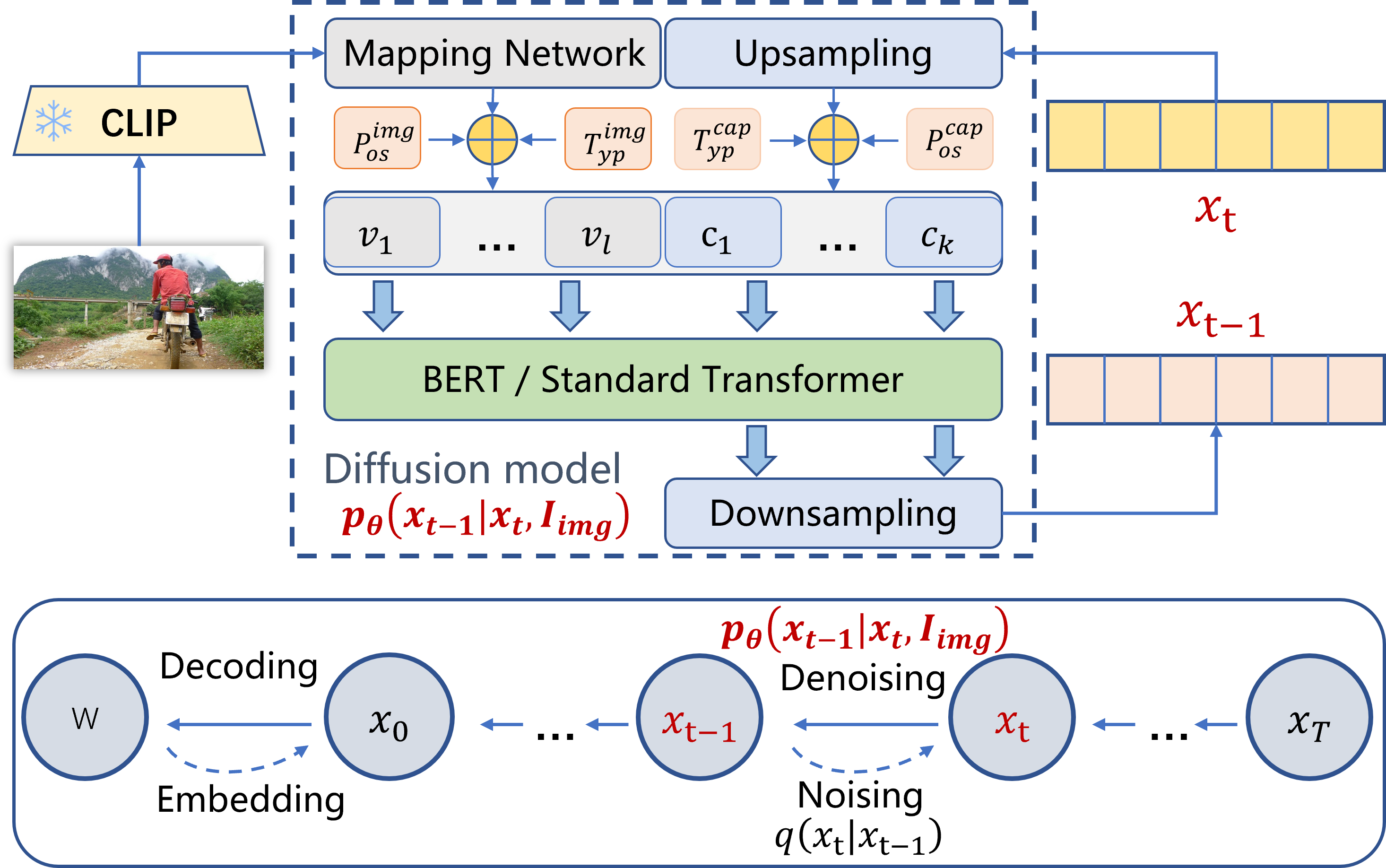}
		\end{center}
		\caption{Illustration of Prefix-diffusion. The bottom lies the diffusion process. The reverse process is defined by $p_{\theta}\left(x_{t-1} \mid x_t, I_{img}\right)$ and the diffusion model is depicted in the upper dashed box. We use the frozen CLIP to extract image features and train a lightweight mapping network to connect the image space and the text space.}
		\label{pic2}
	\end{figure*}

	Recently, the popular diffusion model \citep{sohl2015deep}, which generates samples through an iterative denoising process, has provided a promising path to generate tokens in parallel and inherently increase the diversity of captions.
	Diffusion models \citep{sohl2015deep} have become an active area of research owing to their ability to generate comparable results with GANs \citep{goodfellow2020generative} on computer vision tasks. The strength of diffusion models trained on vast image databases has led to an almost ubiquitous fascination among researchers in producing highly typical content, such as image generation and editing \citep{nichol2021glide, balaji2022ediffi, kim2022diffusionclip, gal2022image}. 
	Nevertheless, the path is blocked by the discreteness of texts and the gap between different modals.

	For the continuous diffusion models \citep{ho2020denoising, nichol2021improved, song2020denoising}, they only work on continuous data but yield inferior results in generating text and image captioning, especially compared to the results of the autoregressive models. 
	To effectively benefit from continuous diffusion, Diffusion-LM \citep{li2022diffusion} extends the the standard diffusion process with an embedding step followed by a rounding step, generating the high-quality text under six control targets. The discreteness of texts has been overcome, whereas the gap between different modals stays unsolved. For image captioning with continuous diffusion, it is a more challenging task, which further requires the fusion of the image information. 
	
	In this paper, we propose a lightweight captioning model based on the continuous diffusion, namely Prefix-diffusion. The model tackles three key problems in image caption generation. Firstly, we utilize diffusion models to solve the limited diversity of the generated captions.
	Noticing that diffusion models have the powerful generative capabilities but few research applied them to image captioning. 
	Secondly, different from image captioning models that have a large number of parameters and are computationally expensive, our framework saves computing resources with the pre-trained CLIP model to extract image features. Last but not least, our method is able to generate more accurate captions in parallel, since it injects prefix image embeddings into the denoising process of the diffusion model. This essentially solves the problem of sequential error accumulation.
	
	Figure \ref{pic1} shows the captions generated by Prefix-diffusion, where the captions accurately describe the content of the image with fluency. Different from the method of beam search, our method can cover all distributions of the training datasets and generate diverse captions.
	
	The overall contributions of our work are:
	\begin{itemize}
		\item We propose a lightweight method Prefix-diffusion to generate diverse captions. Our work tackles the multi-modal issue for the diffusion model and paves the way for scaling it up for image captioning.
		\item Prefix-diffusion generates diverse captions in a variety of forms, which is specifically reflected in the increase of Dist-3 and vocabulary usage by 6.3 and 3.1 compared with the baselines, respectively.
		\item Prefix-diffusion reduces more than 38\% trainable parameters compared with existing CLIP-based methods\citep{nukrai2022text,mokady2021clipcap}, while achieving comparable or even better results in newer metrics. 
		
	\end{itemize}
	
	\section{Related Work}
	
	\subsection{Image Captioning}
	
	The autoregressive models achieve promising performance on image captioning. The next token of the caption is conditioned on the former tokens. To generate more neural captions, \citep{2018Neural}  predicts the slot locations that are explicitly tied to image regions. GET \citep{ji2021improving} captures a more comprehensive global representation by using a novel transformer architecture, to guide the caption generation. Similarly, \citep{li2019entangled, luo2021dual} use transformer to leverage the image information efficiently. Thanks to the powerful multi-modal representation capability of CLIP \citep{radford2021learning}, \citep{mokady2021clipcap, galatolo2021generating} take an image embedding as the input which is encoded by the CLIP visual encoder. Then they use the GPT-2 \citep{radford2019language} model to produce a sequence of words that describe the content of the input image. But autoregressive models suffer from the limitation of generation speed and the accumulation of errors.
	
	Non-autoregressive models have recently attracted attention due to their fast inference speed and generation quality. \citep{gao2019masked} randomly masks the input sequences with certain ratios to train a masked language model, and generates captions parallelly during inference. Considering non-autoregressive image captioning as a cooperative multi-agent problem, \citep{guo2020non} proposes a novel counterfactuals-critical multi-agent learning algorithm to improved the inference speed. \citep{fei2020iterative} proposes a non-autoregressive image captioning approach based on the idea of iterative back modification, which refines the output in a limited number of steps. To determine the length of the image caption, \citep{deng2020length} designs a non-autoregressive decoder for length-controllable image captioning.

	\subsection{Diffusion Model}
	
	Diffusion models \citep{sohl2015deep} have demonstrated impressive capabilities in creative applications. For text-to-image generation, a task of generating a corresponding image from a description, \citep{balaji2022ediffi, nichol2021glide, rombach2022high, gu2022vector} apply discrete diffusion models to produce high-resolution images conditioned on the text prompts. Diffsound \citep{yang2022diffsound} proposes a novel decoder based on the diffusion model to generate high-quality sound. Similarly, ProDiff \citep{huang2022prodiff} studies on diffusion parameterization for text-to-speech and achieves superior sample quality and diversity. In the text generation domain, Diffusion-LM \citep{li2022diffusion} starts with a sequence of Gaussian noise vectors and denoises them incrementally into vectors corresponding to words. Diffusion-LM enables efficient gradient-based methods for controllable generation, achieving promising results in the new forms of complex fine-grained control tasks. Moreover, \citep{ShansanGong2022DiffuSeqST, RobinStrudel2022SelfconditionedED} extend vanilla diffusion models to learn conditional text generation. 
	However, few research applies the diffusion model to image captioning, because of the cross-modal challenge and the discreteness of texts.

	\section{Methodology}
	
	As illustrated in Figure \ref{pic2}, we propose Prefix-diffusion for injecting image features to learn image captioning. Different from image generating, our method requires to map discrete texts to a continuous space by a word embedding. For the conditioned image, we first extract its features by the CLIP image encoder, and then input them to the mapping network to obtain the prefix image embeddings. We then concatenate the prefix image embeddings and the caption embeddings in the denoising process of the diffusion model. The concatenated vectors are fed into a deep neural network (e.g. BERT\citep{kenton2019bert} or the standard transformer). Since our work merely trains a mapping network and a neural network, the trainable parameter scale is reduced significantly.
	
	\textbf{Forward process.} Following Diffusion-LM \citep{li2022diffusion}, we adopt an embedding function $EMB(W)$ to map a discrete word into a continuous space. Define a caption $W$ with $k$ words. Through the embedding function, we have $EMB(W) = \left [ EMB(\omega_{1}), ... , EMB(\omega_{k})\right ]  \in \mathbb{R} ^{k\times d_1}$, where $d_1$ is the dimension of the vector. In our experiments, we find that the value of $d_1$  works well at 48.  Reducing the dimension will decrease the performance, while increasing the dimension will enlarges the computational burden.
	
	For the forward process, diffusion models \citep{ho2020denoising,nichol2021improved,song2020denoising} add noise progressively to training a sample according to a variance schedule $\beta _{1} ,...,\beta _{T}$.
	The forward process has no learnable parameters and we get $x_{t}$ by the following equation:
	\begin{equation}
		x_{t}=\sqrt{1-\beta_{t}} x_{t-1} + \sqrt{\beta _{t} } \epsilon 
	\end{equation}
	where $\epsilon \sim N\left ( 0,1 \right ) $ and $\beta _{t}:0.01\longrightarrow 0.03$ are hyperparameters representing the variance schedule across diffusion steps. We have tried different noise methods, with the truncation linear noise schedule method being the best. We validate this observation in section \ref{Ablation}.
	
	\textbf{Reverse process.} The reverse process generates new samples from $x_{T}\sim N\left ( 0,I \right ) $. The data is sampled using the following reverse diffusion process:
	\begin{multline}
		p_{\theta}\left(x_{t-1} \mid x_t, I_{img}\right)= \\
		\mathcal{N}\left(x_{t-1} ; 
		\mu_\theta\left(x_t, I_{img}\right), \sigma \left ( t \right )^{2}   I\right)
	\end{multline}
	where $I_{img}$ denotes the visual information from CLIP.
	
\begin{figure}[t]
		\begin{center}
			\includegraphics[width=0.95\linewidth]{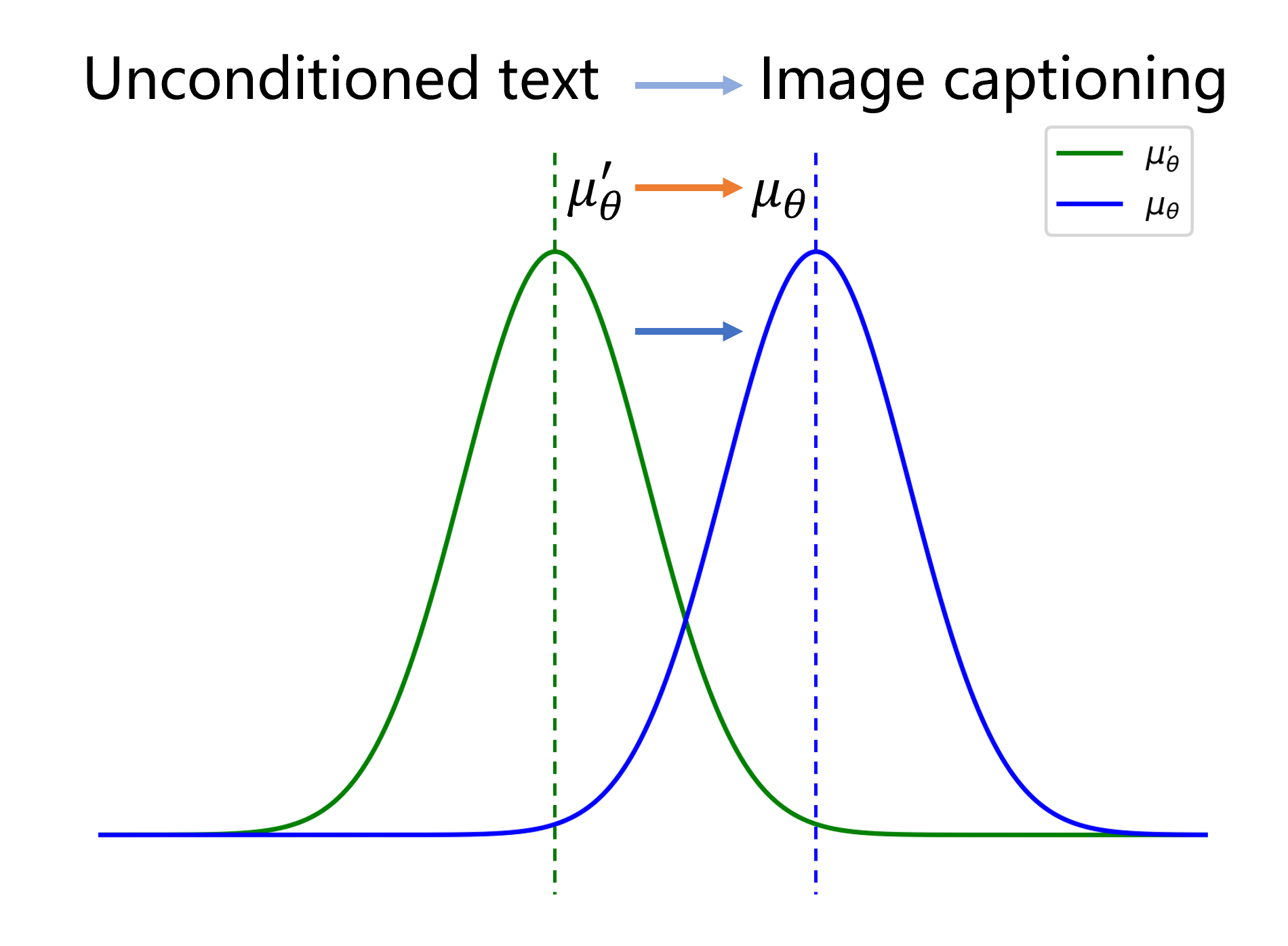}
		\end{center}
		\caption{After we concatenate the image features in the reverse process, the original mean $\mu_\theta^{\prime}$ is changed to $\mu _{\theta }$ in the caption space. 
			Hence, the unconditioned text is converted to an image caption.}
		\label{pic3}
	\end{figure}

	In order to learn the reverse process, neural networks are trained to predict $\mu _{\theta } $ and $\sigma \left ( t \right )^{2}$ is a fixed variance.
	\begin{equation}
		\mu_\theta \left(x_t, I_{img}\right)=\frac{\sqrt{\alpha_t}\left(1-\bar{\alpha}_{t-1}\right)}{1-\bar{\alpha}_t} x_t+\frac{\sqrt{\bar{\alpha}_{t-1}} \beta_t}{1-\bar{\alpha}_t} x_0
	\end{equation}
	\begin{equation}
		\sigma(t)^2=\frac{1-\bar{\alpha}_{t-1}}{1-\bar{\alpha}_t} \beta_t.
	\end{equation}
	In order to get $\mu _{\theta }$, we compute $x_0$ with the following equation:
	\begin{equation}
		x_{0} =\frac{1}{\sqrt{\bar{\alpha _{t}} } } \left ( x_{t}-\sqrt{1-\bar{\alpha _{t}}}\tilde{z}    \right ) 
	\end{equation}
	where $\tilde{z} $ can be obtained by deep neural networks (e.g. transformer).
	\begin{equation}
		\tilde{z} =\Phi (x_{t},I_{img},t) .
	\end{equation}
	Here $\Phi$ denotes the neural network which is depicted in the dashed box in the Figure \ref{pic2}. Since the transformer architecture has been shown to outperform many other architectures on a wide range of text generation tasks, we explored two different transformer architectures as the neural network: BERT and the standard transformer. Different from other continuous diffusion approaches, we inject image features into the transformer architectures. This process changes the original mean in the caption space, as illustrated in Figure \ref{pic3}.
	
	In the following, we will explain in detail how to inject the image information into the model. Firstly we use CLIP to encode image and receive its image features $I_{img}^{'}$. Then we train a mapping network $F$ on $I_{img}^{'}$ and obtain the visual prefix $I_{img}^{m}$ of length $l$:
	\begin{equation}
		\left\{
		\begin{array}{l}
			I_{img}=CLIP(image) \\  I_{img}^{m}=\left \{ v_1^{'},v_2^{'},...,v_l^{'} \right \}=F(I_{img})
		\end{array}
		\right..
	\end{equation}
	
	\begin{table*}[htb]
		\begin{center}
			\resizebox{.97\textwidth}{!}{
				\begin{tabular}{clcccccccccccc}
					\toprule
					\multicolumn{2}{c}{\multirow{2}{*}{\textbf{Method}}} & \multicolumn{6}{c}{\textbf{Common Metrics $ \uparrow $}}        &
					\multicolumn{3}{c}{\textbf{Similarity Score $ \uparrow $}}      &  
					\multicolumn{3}{c}{\textbf{Diversity $ \uparrow $}}  \\ \cmidrule(r){3-8} \cmidrule(r){9-11}   \cmidrule(r){12-14}  
					\multicolumn{2}{c}{}  & B@1    &B@3    & M       & R-L        & C       & S     & CLIP-S   & Ref-CLIP  & P-Bert    & D@2    & D@3    & Voc-u  \\ \midrule
					\multicolumn{2}{c}{MTIC}   & 80.8   & 50.9   & 29.2  & 58.6 & 131.2 &22.6 &60.3 &68.6 &94.0 &7.9 & 16.3 & 8.3\\ 
					\multicolumn{2}{c}{DLCT}  & 81.1   & 51.1  & 29.4 & 58.9 & 133.1 & 22.8 & 60.6 &69.0 &94.1 & 8.1 & 17.1 & 8.3\\ \midrule
					
					\multicolumn{1}{c|}{\multirow{4}{*}{\begin{tabular}[c]{@{}c@{}}Frozen\\Clip\\ Feature\end{tabular}}} &CapDec  & 68.3  & 36.6  & 25.2  & 51.2 &91.7 &18.3 & 60.4 &67.8 &93.4 &8.3 & 14.9 & 1.9 \\
					\multicolumn{1}{c|}{} &ClipCap  & 73.6   & 42.3   & \textbf{26.7}  & 54.4 &105.8 &\underline{19.8} & 60.8 &68.6 &\textbf{93.8} &\underline{11.3} & 21.7 & 2.6 \\
					\multicolumn{1}{c|}{} &Ours(T) & \underline{77.7}     & \underline{43.4}    & 25.8  & \underline{55.8} & \underline{106.3} &19.4  &\underline{63.4} &\underline{70.9} &93.2 &11.2 & \underline{25.9} & \underline{4.7} \\
					\multicolumn{1}{c|}{} &Ours(B)  & \textbf{78.1}    & \textbf{44.2}    & \underline{26.6}  & \textbf{56.1} & \textbf{109.3} &\textbf{20.4}  &\textbf{63.7} &\textbf{71.2} &\underline{93.7} &\textbf{12.7} & \textbf{28.0} & \textbf{5.7}             \\ 
					\bottomrule
			\end{tabular}}
			\caption{The results of image captioning on COCO. For all the metrics, the higher the better. We use boldface to indicate the best performance. The second best result is underlined. Ours(T) and  Ours(B) use a standard transformer and BERT respectively. The values of vocabulary usage are reported at percentage (\%).}
			\label{my-table1}
		\end{center}
	\end{table*}
	
	\begin{table*}[tb]
		\begin{center}
			\resizebox{.97\textwidth}{!}{
				\begin{tabular}{lcccccccccccc}
					\toprule
					\multicolumn{1}{c}{\multirow{2}{*}{\textbf{Method}}} & \multicolumn{6}{c}{\textbf{Common Metrics $ \uparrow $}}        &
					\multicolumn{3}{c}{\textbf{Similarity Score $ \uparrow $}}      &  
					\multicolumn{3}{c}{\textbf{Diversity $ \uparrow $}}  \\ \cmidrule(r){2-7} \cmidrule(r){8-10}   \cmidrule(r){11-13}  
					\multicolumn{1}{c}{}  & B@1    &B@3    & M       & R-L        & C       & S     & CLIP-S   & Ref-CLIP  & P-Bert    & D@2    & D@3    & Voc-u  \\ \midrule
					CapDec  & 57.6    & 27.9   & 20.0  & 44.5 & 42.0 &14.3 &58.0 &61.4 &\underline{92.8} &15.5 & 25.2 & 1.3 \\
					ClipCap  & 67.0    & \underline{35.2}   & \textbf{22.5}  & \underline{49.0} & \underline{60.8} &\textbf{16.5} &60.9 &65.0 &\textbf{93.0} &20.9 & 34.5 & 1.77 \\
					Ours(T) & \underline{68.7}    & 34.9    & 20.1  & 48.7 & 53.8 &14.2 &\underline{61.6} &\underline{66.3} &92.2 &\underline{23.1} & \underline{41.0} & \underline{3.6} \\
					Ours(B)  & \textbf{71.0}    & \textbf{36.2}   & \underline{21.1}  & \textbf{49.3} & \textbf{61.4} & \underline{15.2} &\textbf{64.7} &\textbf{68.6} &92.0 &\textbf{27.6} & \textbf{46.0} & \textbf{4.0}             \\ 
					\bottomrule
			\end{tabular}}
			\caption{The results of image captioning on Flickr30k. For all the metrics, the higher the better. We use boldface to indicate the best performance. The second best result is underlined. }
			\label{my-table2}
		\end{center}
	\end{table*}
	
	We specifically formulate $I_{img}^{m} \in {\mathbb{R}^{l \times d_2}}$ as $\left \{ v_1^{'},v_2^{'},...,v_l^{'} \right \}$ for the convenience of subsequent expression. To save the computation cost, we employ a simple Multi-Layer Perceptron (MLP) as the mapping network. Through an upsampling network, a sequence embedding $x_t$ has the same dimension as $I_{img}^{m}$, denoted as $\left \{ c_1^{'},c_2^{'},...,c_k^{'} \right \}  \in {\mathbb{R}^{k \times d_2}}$. $k$ is the length of the caption and $d_2$ is the dimension of the embedding. 
	
	Before concatenating the visual prefix embedding and the caption embedding, we add positional embedding $P_{os}$ and type embedding $T_{yp}$ to it:
	\begin{equation}
		\begin{split}
			\{ c_1,c_2,...,c_k &\} = \\ 
			\{ &c_1^{'},c_2^{'},...,c_k^{'}  \}
			+P_{os}^{cap} + T_{yp}^{cap} 
		\end{split}
	\end{equation}
	\begin{equation}  
		\begin{split}
			\{ v_1,v_2,...,v_l & \} = \\ 
			\{ &v_1^{'},v_2^{'},...,v_l^{'} \} 
			+P_{os}^{img} + T_{yp}^{img}.
		\end{split}
	\end{equation}
	The positional embedding indicates the model where the feature is located, which is essential information. Similarly, the type embedding tells the model where the image features lie. Then the visual prefix and the caption embedding are concatenated into a sequence $\{ v_1,...,v_l, t_1,...,t_k \}$, and processed by a standard transformer or BERT network:
	\begin{equation}
		\begin{split}
			\{ y_1,y_2,...,y_l, y_{l+1},...,y_{l+k} \} = &\\
			Network(concat( v_1, ..., v_l, &c_1,...,c_k )).
		\end{split}
	\end{equation}
	We split $y_i$ and use $\{y_{l+1},...,y_{l+k} \}$ as the input of the downsampling, yielding the output $x_{t-1} \in {\mathbb{R}^{k \times d_1}}$ of the diffusion model.

	\textbf{Decoding process.} In the decoding process, we strengthen the similarity of images and captions with CLIP scores. The benefit of CLIP in the current work is that it can provide a cosine similarity score between numerous texts and an image. Utilizing the CLIP embedding of an image, we calculate the cosine similarity between the image and the $n$ candidate captions. We then choose the most relevant captions. The similarity is computed as follows:
	\begin{equation}
		similarity\left(I_{img}, W_{txt}^{n} \right)=\frac{I_{img} \cdot W_{txt}^{n}}{\left|I_{img }\right| \cdot\left|W_{txt}^{n}\right|}
	\end{equation}
	where $I_{img}$ is the image features extracted by CLIP and $W_{txt}^{n}$ is the features of the $n$ candidate captions. This is a retrieval-base \citep{ramos2023smallcap, zhao2020image} technique that picks the best appropriate caption from a set of candidate captions. We use this approach based on the advantage of Prefix-diffusion: our model can generate diverse captions with different Gaussian noises. We verify the effectiveness of this retrieval-base method in section \ref{Ablation}. 
	
	\section{Experiment}
	In this section, we conduct quantitative and qualitative experiments to evaluate our approach. We first introduce the implementation details in subsection \ref{Datasets} and \ref{Baseline}. Then we compare the performance of our approach with the others on various evaluation metrics (subsection \ref{Image captioning} and \ref{Cross-domain captioning}). Finally, the ablation experiments (subsection \ref{Ablation}) are also presented to analyze the significance of our design.
	
	\subsection{Dataset and Evaluation Metric}
	\label{Datasets}
	We use COCO \citep{lin2014microsoft} and Flickr30k \citep{plummer2015flickr30k} as the datasets for image captioning. We split the datasets for training, validation, and test according to the Karpathy et al \citep{karpathy2015deep}, where the test sets of the two datasets contain 5000 images and 1000 images respectively. To evaluate the generalization ability of our model, we train the model on one dataset while evaluating on the other.
	
	In this paper, we adopt automatic evaluation to appraise the generated captions. In addition to the common metrics and similarity, we consider two metrics to evaluate the diversity of the generated captions.
	\begin{itemize}
		\item \textit{Common Metrics}. Following the common practice in the literatures, we perform evaluation using BLEU(B@N)\citep{papineni2002bleu}, METEOR(M)\citep{denkowski2014meteor}, ROUGE-L(R-L)\citep{lin2004automatic}, CIDEr(C)\citep{vedantam2015cider}, SPICE(S)\citep{anderson2016spice}. 
		\item \textit{Similarity}. We evaluate the generation by newer metrics: CLIP-S and RefCLIPScore (Ref-CLIP)\citep{hessel2021clipscore} , BERTScore (P-Bert)\citep{zhangbertscore}, which achieve higher correlation with human judgmens.
		\item \textit{Diversity}. Diversity \citep{li2016diversity} is a metric that evaluates the diversity of the generated captions. We report Dist-2(D@2) and Dist-3(D@3) by measuring the diversity of  bigrams and trigrams in the generation. 
		\item \textit{Vocabulary usage}. To analyze the diversity of the generated captions, according to \citep{dai2018neural}, we compute vocabulary usage(Voc-u), which accounts for the percentage of words in the vocabulary that are used in the generated captions.
	\end{itemize}
	
	\begin{figure}[t]
		\begin{center}
			\includegraphics[width=0.95\linewidth]{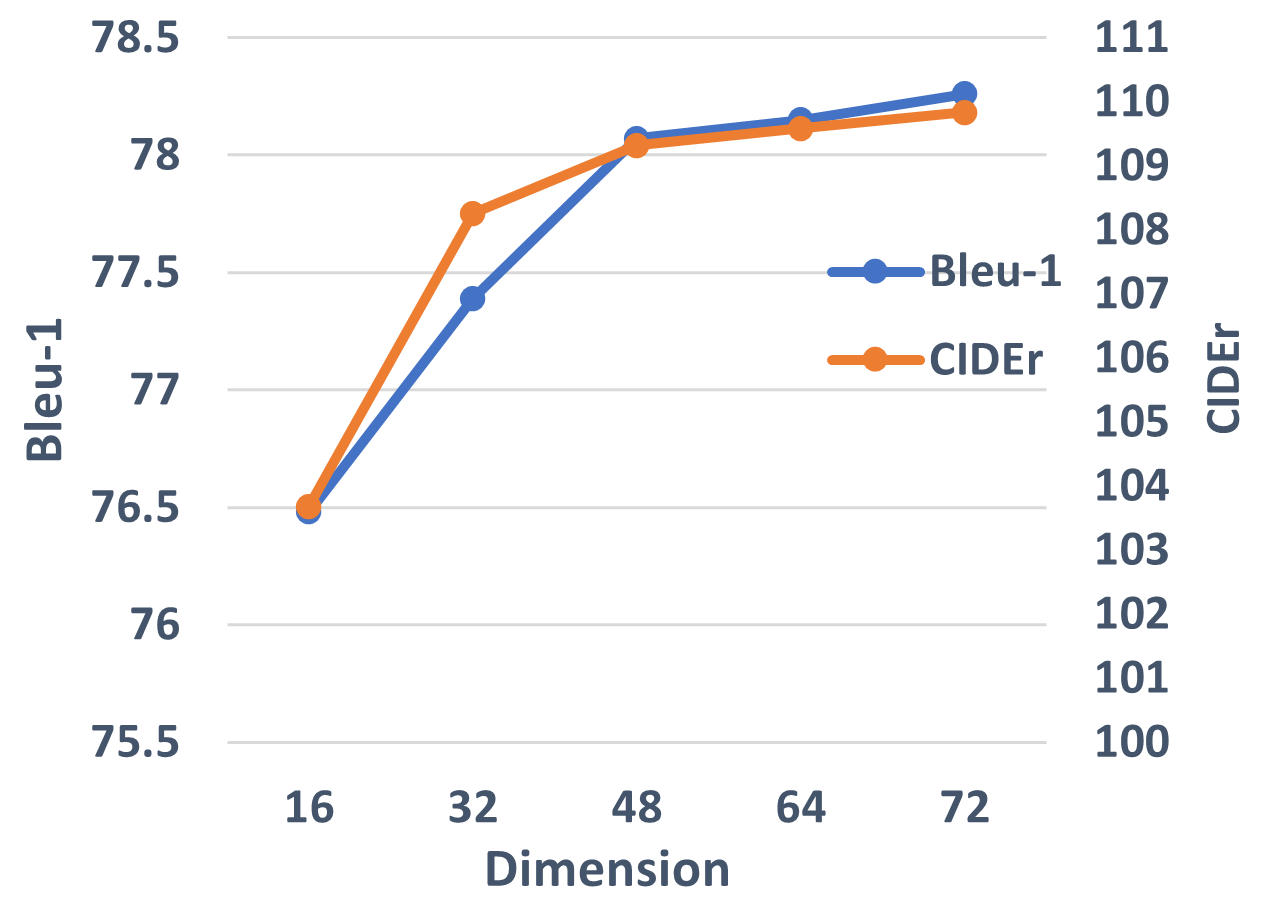}
		\end{center}
		\caption{The performance effect of the word dimension on COCO. We report the metrics of Bleu-1 and CIDEr.}
		\label{pic4}
	\end{figure}
	
	\subsection{Baseline}
	\label{Baseline}
	We adopt the previous competitive image captioning approaches to serve as the baseline models:
	
	\textbf{MTIC} \citep{cornia2020meshed}: MITC is a transformer-based architecture for image captioning. Its image features extracted are by ResNet (denoted as grid-based features).
	
	\textbf{DLCT} \citep{luo2021dual}: DLCT achieves the complementarity of region and grid features for image captioning. To extract visual features, DLCT uses the pretrained Faster-RCNN \citep{ren2015faster}.
	
	\textbf{CapDec} \citep{nukrai2022text}: CapDec is a simple and intuitive approach to learning a captioning model based on CLIP.
	
	\textbf{ClipCap} \citep{mokady2021clipcap}: ClipCap leverages powerful vision-language pre-trained models (CLIP) to simplify the captioning process. And we utilize the MLP mapping network and fine-tunes the language model. All the hyper-parameters are set following its original paper.
	
	Since CapDec and ClipCap use CLIP to extract the same image features and freeze CLIP as our model, we use these methods as the primary baselines. We train our model for 200000 steps, with a batch size of 128. The dimension of word embedding is set to 48 and the diffusion steps $T=1000$. All the experiments are run on NVIDIA Tesla V100 GPUs. In the decoding process, we configure the value of the candidate sentences with $n=5$. Specifically, during the evaluation, we set the denoising steps $T=50$, which greatly reduces the generation time. 
	
	\begin{table}[t]
		\begin{center}
			\resizebox{.97\linewidth}{!}{
				\begin{tabular}{lcccc}
					\toprule
					\multicolumn{1}{c}{\multirow{2}{*}{\textbf{Method}}} & \multicolumn{3}{c}{Human Evaluation$\uparrow$} &\multicolumn{1}{c}{ \multirow{2}{*}{Paras (M) $\downarrow$} } \\  \cmidrule{2-4}
					\multicolumn{1}{c}{} & Fluency       & Sim      & Div   \\ \midrule
					MTIC   & 3.65      & 3.63      & 3.52 & 38.44           \\
					DLCT   & 3.70       & 3.25      & 3.43 & 63.04  	         \\ \midrule
					Capdec   & 3.53       & 2.95      & 3.29 & 178.03  	          \\
					ClipCap    & 3.83       & 3.38      & 3.67     & 155.91   	         \\
					Ours(T)    & 3.79       & 3.84      & 3.95   & \textbf{38.25}    	         \\
					Ours(B)    & \textbf{4.07}       & \textbf{3.95}      & \textbf{4.12}   & 94.83    	          \\ 
					\bottomrule
				\end{tabular}
			}
			\caption{Thr results of human evaluation and the number of trainable parameters for different methods.}
			\label{my-table5}
		\end{center}
	\end{table}
	
	\subsection{Results}
	\label{results}
	
	\subsubsection{Image Captioning}
	\label{Image captioning}
	
	\begin{table*}[htb]
		\begin{center}
			\resizebox{.97\textwidth}{!}{
				\begin{tabular}{lcccccccccccc}
					\toprule
					
					\multicolumn{1}{c}{\multirow{2}{*}{\textbf{Method}}} & \multicolumn{6}{c}{\textbf{Common Metrics $ \uparrow $}}        &
					\multicolumn{3}{c}{\textbf{Similarity Score $ \uparrow $}}      &  
					\multicolumn{3}{c}{\textbf{Diversity $ \uparrow $}}  \\ \cmidrule(r){2-7} \cmidrule(r){8-10}   \cmidrule(r){11-13} 
					\multicolumn{1}{c}{}  & B@1    &B@3    & M       & R-L        & C       & S     & CLIP-S   & Ref-CLIP  & P-Bert    & D@2    & D@3    & Voc-u  \\ \midrule
					\multicolumn{13}{c}{\textit {COCO$\Longrightarrow  $ Flickr30k}}   \\ \midrule
					CapDec  & 57.2   & 23.9   & 17.1  & 40.3 & 30.3 &10.8 &54.4 &58.7 &92.1 &18.5 & 29.4 & 1.2 \\
					ClipCap  & 64.6   & 29.3   & 18.9  & 44.3 & 44.4 &12.5  &56.5 &61.2 & \textbf{92.5} &\textbf{19.7} & 32.7 & 1.3 \\
					Ours(B) & \textbf{69.5}    & \textbf{31.2}    & \textbf{19.3}  & \textbf{46.6} & \textbf{46.8} &\textbf{13.0} &\textbf{61.2} &\textbf{65.3} &91.9 &19.4 & \textbf{37.0} & \textbf{3.0} \\
					\midrule
					\multicolumn{13}{c}{\textit {Flickr30k$\Longrightarrow  $ COCO}}   \\ \midrule
					CapDec  & 44.1    & 15.2   & 15.7  & 36.4 & 25.7 &8.6 &47.7 &51.4 &90.4 &5.5 & 10.4 & 2.0 \\
					ClipCap  & 55.7   & \textbf{23.5}   & \textbf{19.2}  & 42.0 & \textbf{51.3} &\textbf{12.2} &54.9 &60.0 &\textbf{91.1} &11.3 & 21.3 & 3.5 \\
					Ours(B) & \textbf{57.2}   & 22.4    & 17.5  & \textbf{42.5} & 49.3 &11.3 &\textbf{57.5} &\textbf{62.8} &90.4 &\textbf{13.6} & \textbf{29.9} & \textbf{6.6} \\
					\bottomrule
			\end{tabular}}
			\caption{The results of cross-domain captioning. COCO$\Longrightarrow  $ Flickr30k means model trained on COCO while evaluated on Flickr30k, and so is Flickr30k$\Longrightarrow  $ COCO. We use boldface to indicate the best performance.}
			\label{my-table3}
		\end{center}
	\end{table*}
	
	We compare Prefix-diffusion to several baselines with different evaluation metrics, as is shown in Table \ref{my-table1}. Our model outperforms all baselines on CLIP-S and Ref-CLIP metrics, and achieves comparable results on P-Bert score, indicating that the effectiveness of the continuous diffusion on image captioning.  Not only that, we have a significant improvement on some diversity metrics (such as the D@2 and D@3). Furthermore, Prefix-diffusion covers the largest percentage of words, observed from the vocabulary used to generate captions. It implies that captions generated by Prefix-diffusion contain diverse wordings and rich expressions. Our model can generate high-quality captions compared with captioning approaches that extract image feature with CLIP. Prefix-diffusion performs worse than MTIC and DLCT (who not use freeze features for image captioning) on the common metrics, partially due to the proven limitations of word-overlapping-based metrics across various domains\citep{hessel2021clipscore, zhangbertscore}, and also because our generation is more diverse in expression and correctly describe the visual content, which can be observed from similarity score and diversity metrics.

	We also conduct experiments on dataset of Flickr30k, as presented in Table \ref{my-table2}, from which we can draw similar conclusions with the dataset of COCO. Our model achieves impressive performance in the image captioning task compared to the baseline models. In detail, from the results of diversity metrics, we notice that the metrics of Dist-3 and vocabulary usage increase by more than 6.0 and 3.0, respectively. Additionally, we also observe an improvement of 2.6 and 2.8 in CLIP-S and Ref-CLIP metrics, respectively. This indicates that the diffusion model can effectively improve the caption diversity while ensuring coherence and relevance in the generated captions. To generate diverse captions, existing methods tend to generate different captions via top-k sampling. Intuitively, such methods may ignore syntactic diversity and semantic diversity that humans are really interested in. Unlike existing methods, Prefix-diffusion seeks to generate multiple captions with rich expressions from different Gaussian noises.
	
	Figure \ref{pic1} shows the captions generated by Prefix-diffusion. It is observed that the generated captions are pretty consistent with the image as well as keeping the  qualified fluency. Meanwhile, our model is able to generate diverse captions that are more like human-generated.
	
	Furthermore, we conduct human evaluation and report the number of trainable parameters to validate the applicability of our method. As is shown in Table \ref{my-table5}, our model only requires a small number of model parameters. It brings potential advantages of saving memory storage space and computing costs, and thus being much more useful in practice. For human evaluation, we randomly selected 20 samples and presented them in a shuffled manner to 20 annotators.  The annotators rated the fluency, similarity(Sim), and diversity(Div) of the captions on a scale from 1 to 5, with higher scores indicating better quality. From the human evaluation results, We can draw similar conclusions with the automatic evaluation. Our model outperforms the baselines in diversity while holding better fluency and relevance.
	
	The dimension of word embeddings is an important hyper-parameter. The higher dimension leads to more training time and memory usage. To further study the effect of embedding dimension in Prefix-diffusion, we conduct experiments by training with different dimensions. As is shown in Figure \ref{pic4}, the metrics of Bleu-1 and CIDEr are improved as the embedding dimension increases. The reason is that a word embedding becomes richer with semantic information due to the higher dimension. However, there is a performance bottleneck when we continue to increase the dimension of word embeddings. It is observed that the performance trends to be stable when the dimension goes beyond 48.
	
	\begin{table*}[tb]
		\begin{center}
			\resizebox{.97\textwidth}{!}{
				\begin{tabular}{ccccccccccccc}
					\toprule
					\multicolumn{1}{c}{\multirow{2}{*}{\textbf{$n$}}} & \multicolumn{6}{c}{\textbf{Common Metrics $ \uparrow $}}        &
					\multicolumn{3}{c}{\textbf{Similarity Score $ \uparrow $}}      &  
					\multicolumn{3}{c}{\textbf{Diversity $ \uparrow $}}  \\ \cmidrule(r){2-7} \cmidrule(r){8-10}   \cmidrule(r){11-13}  
					\multicolumn{1}{c}{}  & B@1    &B@3    & M       & R-L        & C       & S     & CLIP-S   & Ref-CLIP  & P-Bert    & D@2    & D@3    & Voc-u  \\ \midrule
					1  & 77.2    & 43.6   & 26.0  & 55.6 & 105.2 &19.5 &60.4 &68.6 & 93.1 &11.9 & 26.4 & 5.4 \\
					5  & 78.1    & \textbf{44.2}   & \textbf{26.6}  & \textbf{56.1} & \textbf{109.3} &\textbf{20.4} &63.7 &71.2 &\textbf{93.7} &12.7 & 28.0 & 5.7             \\ 
					10 & \textbf{78.3}    & 43.8    & \textbf{26.6}  & 56.0 & 109.1 &20.3 &65.3 &72.2 &93.4 &13.1 & 28.8 & 5.8 \\
					15  & 78.2    & 43.4    & 26.5  & 55.8 & 108.5 &20.3 &\textbf{66.0} &\textbf{72.6} &93.4 &\textbf{13.4} & \textbf{29.3} & \textbf{5.9}            \\ 
					\bottomrule
			\end{tabular}}
			\caption{The effect of different values of candidate captions. $n=1$ means no cosine similarity calculation in the decoding process.}
			\label{my-table6}
		\end{center}
	\end{table*}
	
	\begin{table}[htb]
		\begin{center}
			\resizebox{.97\linewidth}{!}{
				\begin{tabular}{lcccc}
					\toprule
					\multicolumn{1}{c}{\multirow{2}{*}{\textbf{\begin{tabular}[c]{@{}c@{}}Noise\\ Schedule\end{tabular}}}} & \multicolumn{4}{c}{\textbf{Metrics $ \uparrow $}}           \\ \cmidrule(r){2-5}   
					\multicolumn{1}{c}{}  & B@1    & CLIP-S   & Ref-CLIP  & P-Bert        \\ \midrule
					Square  & 70.5  & \textbf{66.8}  & 72.2  & 92.6    \\
					Linear  & 70.4  & 65.9  & 71.6  & 92.3    \\
					Cosine  & 70.5  & 66.5  & 72.0  & 92.5                \\ 
					T-Cosine  & 72.5  & 66.5  & \textbf{72.3}  & 92.9              \\ 
					T-Linear & \textbf{78.1}  & 63.7  & 71.2  & \textbf{93.7}              \\
					\bottomrule
				\end{tabular}
			}
			\caption{The analysis of different noise schedule in the forward process. T-Linear and T-Cosine means truncation linear noise schedule and truncation cosine noise schedule respectively.}
			\label{my-table7}
		\end{center}
	\end{table}
	
	\subsubsection{Cross-domain Captioning}
	\label{Cross-domain captioning}
	
	We also conduct experiments on cross-domain captioning to evaluate the generalization capability of Prefix-diffusion. The results of the cross-domain evaluation are shown in Table \ref{my-table3}. We train the model on the dataset of a source domain while evaluating it on another dataset. From the results of COCO$\Longrightarrow $Flickr30k, Prefix-diffusion achieves excellent performance over all compared approaches, with the results on the common metrics being the best. In addition, it acquires significant improvements on both Dist-3 and vocabulary usage metrics. This is due to the powerful generative ability of the diffusion model. When we train on flickr30k while evaluating on COCO, the results also show that our approach has strong capability in the cross-domain scenario. By comparing the two results, we find that Prefix-diffusion works even better when trained on a larger dataset, implying the better generalization ability.
	
	\subsubsection{Ablation}
	\label{Ablation}
	
	We perform an ablation study on the dataset of COCO to quantify the contribution of each module in Prefix-diffusion.
	
	Table \ref{my-table6} presents the effect on the number of candidate captions. From the two groups of experiments, $n=1$ and $n=5$, it can be seen that this selection strategy improves the performance of image captioning. We observe a significant increase in the CIDEr metric, which boosts the CIDEr score from 105.2 to 109.3. It confirms the function of calculating the similarity between the image and the candidate captions and choosing the highest. But too many candidate captions lead to a reduction in the performance of the caption fluency. This is because we use the CLIP score as the only similarity selection metric, which may neglect the fluency of captions.
	
	As presented in Table \ref{my-table7}, We investigate the performance of different noise schedules. Observing the results, we conclude that truncated linear noise schedule is able to generate more precise and descriptive captions. We also conclude that the semantic information is corrupted by the complicated noise schedule in the forward process, leading to a more difficult learning problem in the denoising process. 
	
	\section{Conclusion and Future Work}
	
	In this paper, we propose a lightweight network for image captioning in combination with continuous diffusion, called Prefix-diffusion. Experiments and further analysis demonstrate that it can generate diverse captions while maintaining the fluency and relevance of the captions. By trained on one dataset but evaluated on the other, Prefix-diffusion presents remarkable generalization ability. Besides, our model requires a small number of training parameters, which is more applicable in reality. We also conduct ablation experiments to show the effect of the selection strategy and noise schedules. The empirical results verify that Prefix-diffusion has powerful generative ability for image captioning.  For future work, we will continue to explore the potential impact of diffusion models on image captioning. 
	
	\section*{Limitations}
	
	As presented in Table \ref{my-table1} and Table \ref{my-table2}, though Prefix-diffusion can generate diverse captions with relatively less parameters, it is inferior to MTIC and DLCT on the common metrics. But it performs well on newer metrics which have been shown higher correlation with human generation. The reason is that our generated captions have a rich expression that is inconsistent with the reference text, but still convey the same underlying semantics. The length is an important property as it reflects the amount of information carried by a caption. Since our model is a non-autoregressive model, we cannot control the length of the generated text, leading to a less accurate description of the image. We leave this part of exploration for future work.
	
	\section*{Ethics Statement}
	
	Since the proposed Prefix-diffusion can be used to generate captions. With the advantages of being accurate, diverse and descriptive, its generation is more like human-generated. This would benefit image captioning applications on downstream tasks, such as chatting robots and automatic voice guide system. On the other hand, the large number of image captions will make it difficult to distinguish human-wrote from machine-generated. Hence, exploring adversarial attacks on image captioning is necessary. Moreover, excellent captions should involve a variety of words and rich expressions, which prevents them from being too dull or tedious. The diffusion model generates new samples from different noises. Therefore, Prefix-diffusion can be used to improve the diversity of the captions.
	
	\bibliography{anthology,custom}
	\bibliographystyle{acl_natbib}

\end{document}